\documentclass[preprint,12pt]{elsarticle}

\usepackage{amssymb}
\usepackage{amsmath}
\usepackage{graphicx}
\usepackage{float}
\usepackage{xcolor}
\usepackage{booktabs}
\usepackage{silence}
\usepackage{subcaption}
\usepackage[numbers]{natbib}
\usepackage[hidelinks]{hyperref}

\begin{document}

\begin{frontmatter}

%% Title, authors and addresses

%% use the tnoteref command within \title for footnotes;
%% use the tnotetext command for theassociated footnote;
%% use the fnref command within \author or \affiliation for footnotes;
%% use the fntext command for theassociated footnote;
%% use the corref command within \author for corresponding author footnotes;
%% use the cortext command for theassociated footnote;
%% use the ead command for the email address,
%% and the form \ead[url] for the home page:
%% \title{Title\tnoteref{label1}}
%% \tnotetext[label1]{}
%% \author{Name\corref{cor1}\fnref{label2}}
%% \ead{email address}
%% \ead[url]{home page}
%% \fntext[label2]{}
%% \cortext[cor1]{}
%% \affiliation{organization={},
%%             addressline={},
%%             city={},
%%             postcode={},
%%             state={},
%%             country={}}
%% \fntext[label3]{}

\title{Multi-Graph Inductive Representation Learning for Large-Scale Urban Rail Demand Prediction under Disruptions} %% Article title

%% use optional labels to link authors explicitly to addresses:
%% \author[label1,label2]{}
%% \affiliation[label1]{organization={},
%%             addressline={},
%%             city={},
%%             postcode={},
%%             state={},
%%             country={}}
%%
%% \affiliation[label2]{organization={},
%%             addressline={},
%%             city={},
%%             postcode={},
%%             state={},
%%             country={}}

\author[1]{Dang Viet Anh Nguyen} %% Author name
\author[2]{J. Victor Flensburg} %% Author name
\author[3]{Fabrizio Cerreto}
\author[4]{Bianca Pascariu}
\author[4]{Paola Pellegrini}
\author[1]{Carlos Lima Azevedo}
\author[1]{Filipe Rodrigues}

%% Author affiliation
\affiliation[1]{organization={Department of Technology, Management, and Economics, Technical University of Denmark (DTU)},%Department and Organization
            % addressline={}, 
            % city={},
            postcode={2800 Kongens Lyngby}, 
            % state={},
            country={Denmark}}

%% Author affiliation
\affiliation[2]{organization={Traffic Division, Banedanmark},%Department and Organization
            % addressline={}, 
            % city={},
            postcode={4100 Ringsted}, 
            % state={},
            country={Denmark}}

\affiliation[3]{organization={Metroselskabet og Hovedstadens Letbane},%Department and Organization
            % addressline={}, 
            % city={},
            postcode={2300 København S}, 
            % state={},
            country={Denmark}}

\affiliation[4]{organization={COSYS-ESTAS, Gustave Eiffel University},%Department and Organization
            % addressline={}, 
            % city={},
            postcode={F-59650 Villeneuve d’Ascq}, 
            % state={},
            country={France}}

%% Abstract
\begin{abstract}
%% Text of abstract
With the expansion of cities over time, URT (Urban Rail Transit) networks have also grown significantly. Demand prediction plays an important role in supporting planning, scheduling, fleet management, and other operational decisions. In this study, we propose an Origin-Destination (OD) demand prediction model called Multi-Graph Inductive Representation Learning (mGraphSAGE) for large-scale URT networks under operational uncertainties. Our main contributions are twofold. First, we enhance prediction results while ensuring scalability for large networks: we use multiple graphs simultaneously, where each OD pair is represented as a node on a graph, and distinct OD relationships, such as temporal and spatial correlations, are considered. As a second contribution, we demonstrate the importance of incorporating operational uncertainties, such as train delays and cancellations, as inputs in demand prediction for daily operations. The model is validated on three different scales of the URT network in Copenhagen, Denmark. Experimental results show that by leveraging information from neighboring ODs and learning node representations via sampling and aggregation, mGraphSAGE is particularly suitable for OD demand prediction in large-scale URT networks, outperforming reference machine learning methods. Furthermore, during periods with train cancellations and delays, the performance gap between mGraphSAGE and other methods improves, demonstrating its ability to leverage system reliability information for predicting OD demand under system disruptions.
\end{abstract}

% %%Graphical abstract
% \begin{graphicalabstract}
% %\includegraphics{grabs}
% \end{graphicalabstract}

%%Research highlights
% \begin{highlights}
% \item Research highlight 1
% \item Research highlight 2
% \end{highlights}

%% Keywords
\begin{keyword}

OD demand prediction \sep Urban transit rail \sep Graph Neural Networks \sep Large scale

%% keywords here, in the form: keyword \sep keyword

%% PACS codes here, in the form: \PACS code \sep code

%% MSC codes here, in the form: \MSC code \sep code
%% or \MSC[2008] code \sep code (2000 is the default)

\end{keyword}

\end{frontmatter}

%% Add \usepackage{lineno} before \begin{document} and uncomment 
%% following line to enable line numbers
%% \linenumbers

%% main text
%%

%% Use \section commands to start a section

\section{Introduction}
% \textbf{Main text:  7500 words}
In today's urban areas, the development of public transport is critical in addressing the challenges of traffic congestion, carbon emissions, and sustainable mobility. Along with bus systems, urban rail transit (URT) provides transportation for millions of daily trips in cities worldwide. For instance, the S-train system in Copenhagen, Denmark, served 106.179 million passengers in 2023~\cite{Passenge59:online}. Therefore, maintaining a reliable and efficient URT system is essential not only for citizens’ quality of life but also for economic productivity and environmental sustainability. 

In the research field of URT operation, origin-destination (OD) prediction, which estimates the number of passengers traveling between origin and destination stations, is a crucial task to support planning, scheduling, fleet management, and other operational decisions \cite{jiang2022deep}. Unlike zone-based demand forecasting, which has been extensively studied in previous research, OD prediction is more challenging due to the complexities of URT network structures, transfer patterns, passenger travel behaviors, and partial observability \cite{aboah2021identifyingfactorsinfluenceurban}. With the trend of urbanization and the expansion of URT networks in most cities worldwide, the scale of OD networks has grown considerably, raising the need for more scalable and accurate predictive models. Additionally, the demand for the URT system usually changes based on the level of system reliability. Therefore, the URT demand prediction model should account for the system's reliability and adapt to dynamic changes in the network.

With the advancement of machine learning, computing power, and data availability, recent studies have explored advanced techniques in deep learning to capture complex dependencies and specific demand patterns in URT networks for OD prediction \cite{jiang2022deep, zhang2021short, noursalehi2021dynamic}. Most previous studies focus on tackling different aspects of OD demand prediction in URT networks, such as real-time data availability, sparsity, high dimensionality, and ridership quantity relationships. Very few studies address the challenge of predicting OD demand under disrupted scenarios including unexpected events like train cancellations or delays \cite{zhang2024physics}. Additionally, although prediction models based on graph neural networks (GNNs) have been widely adopted in traffic prediction \cite{yu2017spatio, zhao2019t, ke2021predicting, ali2022exploiting}, their application in URT OD demand prediction is still limited. GNNs are known to efficiently extract features with complex topological characteristics. However, one of the challenges related to their utilization is their scalability. Most traditional GNN models rely on spectral decomposition or matrix factorization techniques, which are transductive and do not scale well to large graphs.

Given these considerations, this study proposes a short-term OD demand prediction model that is both scalable for large-scale URT networks and robust under various scenarios of URT system reliability, based on graph inductive representation learning \cite{hamilton2018inductiverepresentationlearninglarge}. Specifically, the proposed method models the OD demand of the URT network as different graphs, establishing connectivity via various temporal and spatial correlations. The proposed graphs are then trained using graph inductive representation learning technique designed for inductively learning node representation in a large graph. By learning through sampling and aggregating features from neighboring ODs, we achieve a prediction model that performs well and scales effectively to large, complex real-world URT networks. Additionally, our model incorporates unique features regarding URT system reliability, enhancing the prediction model's performance under train cancellation or delay scenarios. The main contributions of this paper are summarized as follows:
\begin{itemize}
\item We propose a short-term OD demand prediction model that combines graph inductive representation learning with a multi-graph framework to enhance the performance and scalability of the model in a large-scale URT network. The proposed model is validated using three different scales of the real-world S-train network in Copenhagen, Denmark.
\item By incorporating unique system reliability features, such as train cancellations and delays, we enhance the model's performance across different disruption scenarios in the URT network. Experiments conducted under various conditions of train cancellations and delays further validate the effectiveness of the proposed model.
\end{itemize}

The remainder of this paper is organized as follows: Section \ref{related} highlights previous related work on short-term OD demand prediction in URT. Section \ref{method} details the research problem, graph modeling, feature extraction, and the proposed model - mGraphSAGE. Section \ref{experiment} presents the computational experiments and comparative analysis with reference methods to validate the performance of the proposed model. Finally, Section \ref{conclude} summarizes the main findings, limitations of the proposed model, and suggests possible future research directions.

\section{Related Works}
\label{related}

In the last decade, numerous studies have focused on traffic prediction problems, such as road traffic flow, road origin-destination flow, bike/taxi/ride-hailing flow, and URT inflow/outflow. However, relatively few studies have addressed short-term URT OD demand prediction, primarily due to several challenges that distinguish it from other prediction problems. These challenges include the real-time availability of OD data, data sparsity, and high dimensionality.

The real-time availability of OD data refers to the difficulty in observing OD demand, as each OD demand represents a passenger's travel behavior that can only be collected after the trip is completed \cite{zhang2021short, noursalehi2021dynamic}. Incomplete travel information can adversely affect prediction accuracy, necessitating effective models to obtain nearly complete real-time OD demand data \cite{zhang2024physics}. Data sparsity arises from the low or zero demand for many ODs due to time-dependent travel patterns, the large dimensionality of OD flows, and variations across different city areas \cite{zhang2021short}. High dimensionality refers to the quadratic relationship between the total number of ODs and the number of stations, resulting in significantly larger OD networks in URT compared to other transportation modes.

Recent research has tackled these challenges through various models and leveraged deep learning advancements to improve prediction model performance. Jiang et al.~\cite{jiang2022deep} proposed a deep learning prediction model using temporally shifted graph convolution and LSTM with an attention mechanism to capture temporal and spatial dependencies, while incorporating a reconstruction mechanism to address partial observability of OD flow information. Liu et al. \cite{liu2022online} addressed real-time data availability by developing a deep learning model that jointly learns from heterogeneous information, including incomplete OD matrices, unfinished order vectors, and destination-origin (DO) matrices representing the origin distribution of outgoing ridership demand. Zhang et al.~\cite{zhang2021short} introduced a channel-wise attentive-split convolutional neural network (CNN) with an inflow/outflow-gated mechanism to address real-time data availability by capturing correlations between inflows/outflows and OD flows. Noursalehi et al.~\cite{noursalehi2021dynamic} proposed a multi-resolution spatio-temporal neural network model to capture spatial and temporal dependencies while utilizing exit-based observations to handle partial observability. Zhu et al.~\cite{zhu2023two} introduced a two-stage OD flow prediction model that includes flow prediction and separation rate estimation, with future OD demand derived by combining these two outputs. Cheng et al.~\cite{cheng2022real} proposed a model using Singular Value Decomposition (SVD) and a low-rank high-order vector autoregression model to approximate the OD data and predict task where the problem of OD data availability is tackled by time-evolving features. Recently, Zhang et al. \cite{zhang2024physics} proposed a physics-guided adaptive graph spatial-temporal attention network, utilizing a masked physics-guided loss function to capture relationships between OD demand and inbound flow. This approach also developed real-time OD estimation and OD demand matrix compression to tackle real-time data availability and matrix sparsity issues. 

% Despite the success of Graph Neural Networks (GNNs) in various traffic forecasting problems due to their ability to capture spatial dependencies in non-Euclidean graph structures, their application in URT OD demand prediction remains limited due to the aforementioned challenges. Additionally, the high dimensionality of OD networks challenges traditional GNN methods due to limitations in their scalability to large graphs. 

This study aims to propose a GNN-based method that efficiently addresses the three mentioned challenges in URT OD demand prediction and is capable of handling real-world URT network instances. Additionally, the reliability of the public transportation system plays an important role in ensuring travelers attraction, making it essential to incorporate the impacts of potential disruption scenarios into the prediction model \cite{cats2014dynamic}. Considering the impact of train disruptions in the URT system, while recent research \cite{zhang2024physics} has addressed unexpected factors such as those caused by COVID-19, our study focuses on more general daily operational issues, such as train delays and cancellations. The goal is to develop a prediction model that performs effectively under various operational scenarios of the URT system.

\section{Methodology}
\label{method}

In this section, we present our research problem on OD prediction in URT, describe the features in our data, and introduce the proposed mGraphSAGE model for OD prediction in URT systems.

\subsection{Research problem and features}

%This paper considers short-term OD demand prediction in an URT network, as described in Section \ref{sub:URO}. 
The OD demand prediction task at the network level is challenging due to the complex spatial and temporal dependencies among the demands of different OD pairs. Additionally, the level of observability of passenger trips in the metro at the predicted time step varies because each OD demand can only be recorded after the passenger has completed their trip. This makes some of the OD demand only partially observed at prediction time. To structure the prediction timeline, we divide the prediction horizon into discrete intervals; at the beginning of each interval, the prediction task utilizes historical information to forecast demand for that specific interval.

Let $d^{i}_{t}$ be the number of passengers departing at time interval $t$ and having already completed their trip through OD $i \in V$ (the set of OD pairs) before the prediction time. $p^{o}_{t}$ denotes the number of passengers departing at origin station $o$ of OD $i$ but not yet completed their trip at prediction time. Let $x^{i}_{t} = d^{i}_{t} + p^{o}_{t}$ be the sum of the number of trips completed through OD $i$ and the number of uncompleted trips departing from the origin station of OD $i$. Thus, $\textbf{D}_{t} = \{d^{1}_{t}, d^{2}_{t}, \dots, d^{n}_{t}\}$ is the vector of completed trips of all OD pairs on the graph departing at time interval $t$. $\textbf{P}_{t} = \{p^{1}_{t}, p^{2}_{t}, \dots ,p^{n}_{t}\}$ is the vector of uncompleted trips of all $n$ origin stations of $n$ ODs. $\textbf{X}_{t} = \{x^{1}_{t}, x^{2}_{t}, \dots ,x^{n}_{t}\}$ is the vector of aggregated passenger demand traveling through all OD pairs (the sum of completed trips on an OD pair and passengers departing from the origin station of the respective OD pair) departing at time interval $t$. 

\begin{figure}[H]
    \centering
    \includegraphics[width = \textwidth]{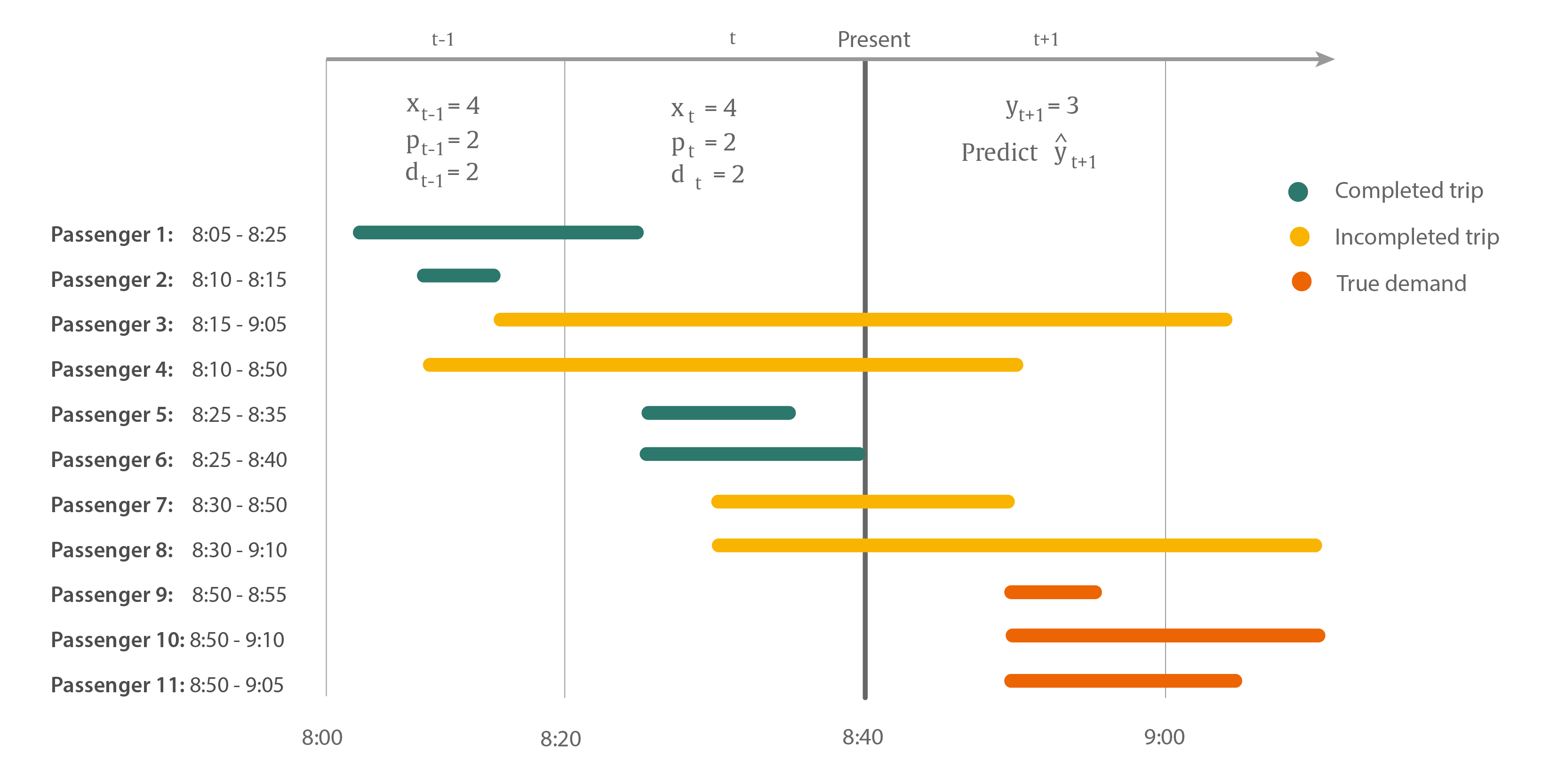}
    \caption{Visualization of demand feature extraction for a given OD and for a given prediction time}
    \label{fig:feat}
\end{figure}

Figure \ref{fig:feat} illustrates how completed trips $\textbf{D}$, uncompleted trips $\textbf{P}$, and aggregated demand $\textbf{X}$ are extracted for each prediction time and for a given OD. For instance, if the prediction time is 8:40, in the interval $t-1$ from 8:00 to 8:20, there are a total of 4 passengers departing, but only 2 of them completed their trip before 8:40. Therefore, we identify $x_{t-1} = 4$, $p_{t-1} = 2$, and $d_{t-1} = 2$. Another thing to note is that $x_{t}$, $p_{t}$, and $d_{t}$ will be updated as the prediction time progresses. For example, when the prediction time moves to 9:00, and the trip of passenger 4 is completed at 8:50, then $d_{t-1} = 3$ and $p_{t-1} = 1$ instead of $2$ and $2$ as previously.Using the information of aggregated demand on each OD, $\textbf{X}$, and completed trips on each OD, $\textbf{D}$, we aim to tackle the partial observability challenges in the OD prediction task.

According to \cite{zhang2017deep, ke2021predicting}, OD demand temporal dependencies have two main types: \textit{tendency} and \textit{periodicity}. The former refers to present demand being affected by previous demand in the last several time intervals, while the latter relates to the repetition of daily and weekly demand patterns. To account for \textit{tendency}, we gather the data of $\textbf{X}$ and $\textbf{D}$ from the last 8 time intervals before the prediction time, specifically $\{\textbf{X}_{t}, \textbf{X}_{t-1}, \dots, \textbf{X}_{t-7}\}$ and $\{\textbf{D}_{t}, \textbf{D}_{t-1}, \dots, \textbf{D}_{t-7}\}$. The choice of the number of previous intervals needs to be determined by empirical research. To capture daily and weekly patterns from \textit{periodicity}, we use the following features:
\begin{itemize}
    \item One-hot encoding features for indicating day of the week: $\textbf{f}_{w}$;
    \item One-hot encoding features for indicating time interval of the day: $\textbf{f}_{t}$.  
\end{itemize}

\noindent We also add one-hot encoded node ID features $\textbf{f}_{id}$ to distinguish different nodes (OD pairs) on the graph. However, in some case using one-hot encoding for each OD pair in the large-scale case would result in a massive increase in feature dimensions. To avoid this, we use a different encoding method to distinguish OD pairs. We add a node-ID feature as a one-dimensional vector with a size equal to the number of stations. The element having index corresponding to the origin station is indicated as 1, while the element having index corresponding to the destination station is indicated as -1.

On top of \textit{tendency} and \textit{periodicity}, we capture \textit{reliability} to account for impact of operations reliability of the URT system in demand. Therefore, we incorporate 12 supply features denoted by $\textbf{f}_{s}$, which measure the system's reliability for each OD pair. According to \cite{kathuria2020review, tirachini2022headway}, popular reliability measures of public transportation systems include waiting time, regularity, and travel time. In the context of our study, regularity refers to the frequency of train arrivals and cancellations within one hour, while waiting time and travel time cover the delays of trains at specific stations relative to the OD pairs. Therefore, we measure the reliability of the system for each OD pair based on three dimensions at both the origin and destination stations as follows:

\begin{itemize}
\item The number of trains at the station during the last hour for each line, with respect to the OD direction;
\item Mean delay of trains at the station during the last hour for each line, with respect to the OD direction;
\item Proportion of trains that have been cancelled at the station during the last hour for each line, with respect to the OD direction.
\end{itemize}

\noindent Because each station has multiple lines, we take the mean and maximum values for each dimension at both the origin and destination stations. Therefore, we have a total of 12 features related to the reliability of the railway system.

Given all the observed aggregated OD demand data and the completed trips from all ODs  and additional information such as date, time, and system reliability up to time interval $t$, our study aims to learn a function $f(\cdot): \mathbb{R}^{N \times T} \rightarrow \mathbb{R}^{N}$ to predict the complete OD demand of the urban rail network for the next time interval $\textbf{C}_{t+1}$:

$$[\textbf{X}_{t-7}, \dots, \textbf{X}_{t}, \textbf{D}_{t-7}, \dots, \textbf{D}_{t}, \textbf{f}_{w}, \textbf{f}_{t}, \textbf{f}_{id}, \textbf{f}_{s}] \xrightarrow{f(\cdot)} \textbf{C}_{t+1}$$

\subsection{OD graph in URT}
\label{sub:URO}

In conventional modeling of the URT network as a graph, stations are represented as vertices, with edges indicating passenger flow between them. While this approach provides a clear and intuitive understanding of the network structure, it does not effectively represent the relationships in demand patterns between OD pairs and can lead to significant graph size increases when considering OD demand as edges. In this study, we address the OD prediction problem in a large-scale URT network by modeling each OD pair as a vertex in the graph. This strategy reduces both the complexity and size of the graph, facilitating more efficient node prediction for OD demand. Specifically, we define an OD graph \( G = (V, E, A) \), where \( V \) is the set of vertices representing OD pairs and \( E \) represents the relationship in demand patterns between these pairs. The adjacency matrix \( A \in \mathbb{R}^{n \times n} \) defines the relationships between the vertices, where \( n \) is the number of vertices (OD pairs) and the element \( A_{ij} \) indicates the presence of an edge between vertices \( i \) and \( j \in V \). 

\subsection{The  mGraphSAGE model}

In this section, we detail the proposed method based on multi-graph inductive representation learning (mGraphSAGE) for predicting OD demand in a large scale URT network. 

% The overall architecture of the proposed mGraphSAGE method for predicting OD demand is illustrated in Figure \ref{fig:model}. 

\subsubsection{Modeling temporal-spatial correlations between OD pairs} \label{sec:adjacency}

A key challenge in applying GNN-based methods for OD demand prediction is defining the connections among the vertices of the graph. In the context of an urban rail network, the connectivity of the graph can be defined in multiple ways based on both temporal and spatial correlations. To show the potential of mGraphSAGE in combining different graph structures, we construct four different graphs with different adjacency matrices: (i) temporal correlation-based graph \(G_{t}(V,E,A^{t})\), (ii) centroid distance graph \(G_{s}(V,E,A^{s})\), (iii) origin distance graph \(G_{o}(V,E,A^{o})\), and (iv) destination distance graph \(G_{d}(V,E,A^{d})\). 

We chose distance-based methods over available cluster-based measures to quantify temporal and spatial correlations between OD pairs because they provide a clearer and more interpretable way to evaluate the relationship between two OD pairs, making it easier to understand why specific nodes are connected in the graph. These methods also facilitate precise comparisons between individual OD pairs when constructing the adjacency matrix, whereas clustering methods tend to generalize relationships into broader groups, potentially overlooking finer details.

\noindent Following are the details of the four different graphs we propose:

(i) \textbf{Temporal correlation-based graph}: This graph's connectivity is determined by measuring the similarity in the demand patterns of the last interval \(t\) between OD pairs. We use Dynamic Time Warping (DTW) to measure the level of similarity of all OD pairs to form the temporal distance matrix \(D^{t}\). Then, we set a threshold \(\sigma_{t}\) to determine whether these pairs of OD will be connected in the graph. The adjacency matrix of the temporal correlation-based graph \(A^{t}\) is constructed as follows:

\begin{align}
    A^{t}_{ij}=
    \begin{cases}
      1, & \text{if}\ D^{t}_{ij} \leq \sigma_{t} \\
      0, & \text{otherwise}
    \end{cases}
  \end{align}

(ii) \textbf{Centroid distance graph}: Inspired by \cite{ke2021predicting}, we measure the geographical distance between origin-destination (OD) pairs based on the distance between the centroids of the origins and the centroids of the destinations for each OD pair. The rationale behind establishing connectivity is that we anticipate OD pairs that are geographically close to exhibit strong demand correlation. Therefore, we construct the centroid distance matrix \(D^{s}\) as below:

\begin{align}
    D^{s}_{ij} = \sqrt{\left( \frac{x^{o}_{j} + x^{d}_{j}}{2} - \frac{x^{o}_{i} + x^{d}_{i}}{2}\right)^2 + \left( \frac{y^{o}_{j} + y^{d}_{j}}{2} - \frac{y^{o}_{i} + y^{d}_{i}}{2} \right)^{2}} \quad \forall i,j \in V
\end{align}
where \(x^{o}_{i}, y^{o}_{i}, x^{d}_{i}, y^{d}_{i}\) are the UTM coordinates of the origin and destination stations of OD pair \(i\), respectively, and \(x^{o}_{j}, y^{o}_{j}, x^{d}_{j}, y^{d}_{j}\) are the UTM coordinates of the origin and destination stations of OD pair \(j\).

From the centroid distance matrix \(D^{s}\), we also set another threshold \(\sigma_{s}\) where closer OD pairs should be connected in the graph. The adjacency matrix of the centroid distance graph is established similarly to the temporal correlation-based graph, where the element \(A^{s}_{ij} = 1\) if \(D^{s}_{ij} \leq \sigma_{s}\), and \(A^{s}_{ij} = 0\) otherwise.

(iii, iv) \textbf{Origin distance graph} and \textbf{destination distance graph}: We also observe that OD pairs with close origins and destinations typically involve transit trips between nearby stations. Therefore, to capture the demand correlation of OD pairs that have close origin and destination distances, we construct two distance matrices \(D^{o}\) and \(D^{d}\) to measure the distances between the origins and destinations of each OD pair respectively:

\begin{align}
    D^{o}_{ij} = \sqrt{ \left( x^{o}_{j} - x^{o}_{i} \right)^2 + \left( y^{o}_{j} - y^{o}_{i} \right)^{2}} \quad \forall i,j \in V
\end{align}
\begin{align}
    D^{d}_{ij} = \sqrt{ \left( x^{d}_{j} - x^{d}_{i} \right)^2 + \left( y^{d}_{j} - y^{d}_{i} \right)^{2}} \quad \forall i,j \in V
\end{align}
We also set thresholds \(\sigma^{o}\) and \(\sigma^{d}\) to define the adjacency matrices \(A^{o}\) and \(A^{d}\) of graphs \(G_{o}\) and \(G_{d}\).

\subsubsection{Multi-graph inductive representation learning (mGraphSAGE)}
\begin{figure}[H]
    \centering
    \includegraphics[width = \textwidth]{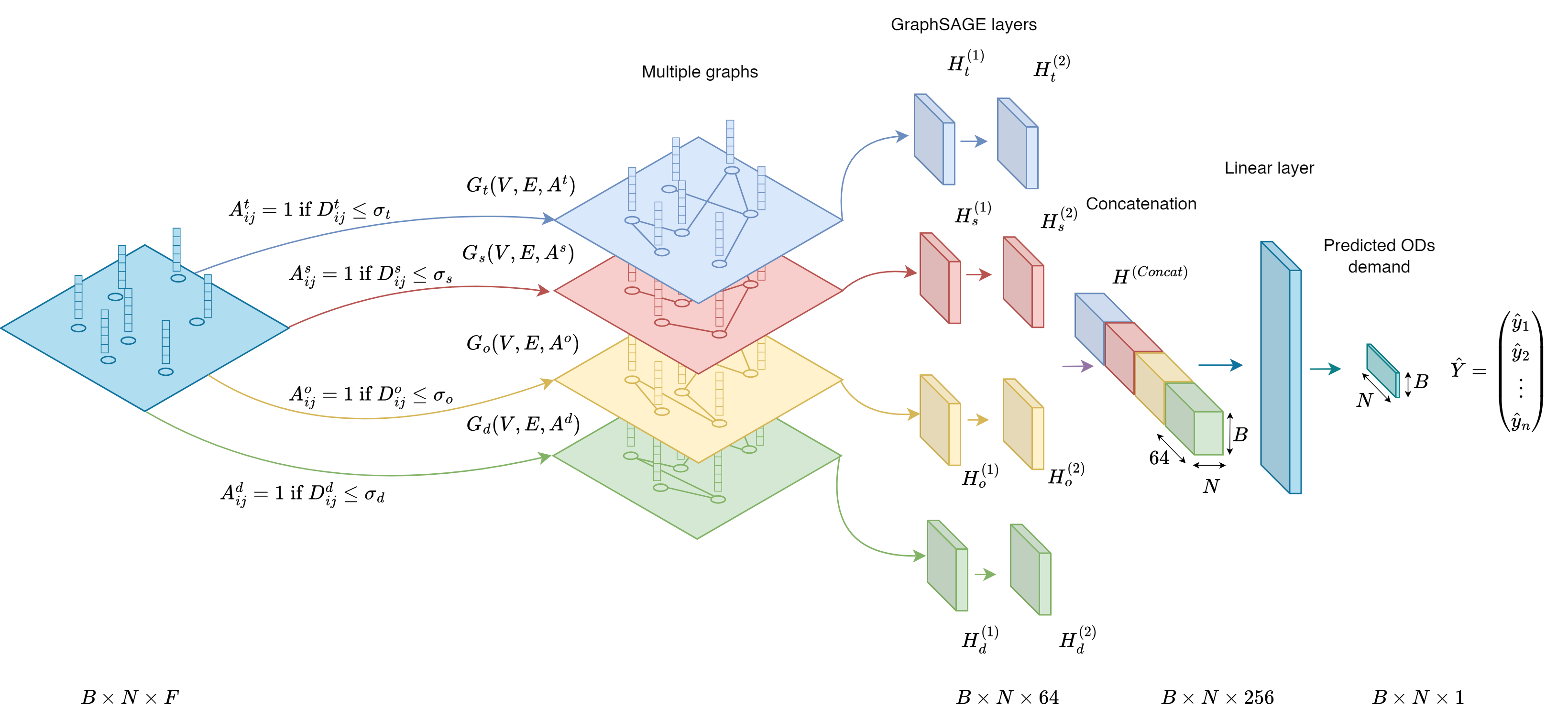}
    \caption{Multi-graph Inductive Representation Learning (mGraphSAGE)}
    \label{fig:model}
\end{figure}

Figure \ref{fig:model} illustrates the architecture of mGraphSAGE, where $F$, $N$, and $B$ denote the feature length, the number of nodes (OD pairs), and the batch size, respectively. We use four adjacency matrices \(A^{t}, A^{s}, A^{o}, A^{d}\) (as discussed in Section \ref{sec:adjacency}) to establish the connections of the graph, forming four graphs \(G_{t}, G_{s}, G_{o}\), and \(G_{d}\). These four graphs are then trained in four separate channels using the GraphSAGE algorithm for learning node representations.

GraphSAGE (Graph Sample and Aggregation) is a method for inductively learning node embeddings in large graphs. First introduced by Hamilton et al. \cite{hamilton2018inductiverepresentationlearninglarge}, this method addresses scalability and generalization issues in traditional graph convolutional networks (GCNs). GraphSAGE is the first inductive learning method that, instead of learning specific embeddings for nodes, learns an aggregation function that can be applied to any node's neighborhood, regardless of whether the node was present during training. This property allows GraphSAGE to generalize to unseen nodes or graphs without the need for retraining. As a result, a prediction model based on GraphSAGE is an efficient way to handle the sparsity in OD demand data within URT networks.

Given a graph \( G = (V, E) \), where \( V \) is the set of nodes and \( E \) is the set of edges, let \( \mathbf{h}_v^{(k)} \) denote the embedding of node \( v \) at the \( k \)-th layer of the model. The GraphSAGE node embedding generation procedure is as follows:
\begin{enumerate}
\item GraphSAGE samples a fixed set of neighbors \( \mathcal{N}(v) \) for each node \( v \), rather than using all neighbors. This sampling helps to efficiently handle large graphs. Once neighbors \( \mathcal{N}(v) \) are sampled, GraphSAGE uses an aggregation function \( \text{AGGREGATE}_{k} \) at layer \( k \) to combine information from these neighboring nodes:
   \[
   \mathbf{h}_{\mathcal{N}(v)}^{(k)} = \text{AGGREGATE}_{k} \left( \left\{ \mathbf{h}_{u}^{(k-1)} \mid u \in \mathcal{N}(v) \right\} \right),
   \]
   where \( \mathbf{h}_{u}^{(k-1)} \) is the embedding of node \( u \) at layer \( k-1 \).
   
\item The aggregated information \( \mathbf{h}_{\mathcal{N}(v)}^{(k)} \) is then used to update the representation of each node \( \mathbf{h}_{v}^{(k)} \). This step, known as node embedding update, combines both the node's own features and the aggregated features from its neighbors. This is done using a fully connected layer with a nonlinear activation function \( \sigma \):
   \[
   \mathbf{h}_{v}^{(k)} = \sigma \left( \mathbf{W}^{(k)} \cdot \text{CONCAT} \left( \mathbf{h}_{v}^{(k-1)}, \mathbf{h}_{\mathcal{N}(v)}^{(k)} \right) \right).
   \]
   
   Here, \( \mathbf{W}^{(k)} \) denotes the weight matrix at layer \( k \), and \( \text{CONCAT}(\cdot) \) denotes the concatenation operation along specified dimensions.
   
\item Different types of aggregation functions can be used, such as mean aggregator, LSTM aggregator, and pooling aggregator. In this study, we use the mean aggregator. Thus, the node representation update using the mean aggregator is given by:
   \[
   \mathbf{h}_{v}^{(k)} \gets \sigma \left( \mathbf{W}^{(k)} \cdot \text{MEAN} \left( \left\{ \mathbf{h}_{v}^{(k-1)} \right\} \cup \left\{ \mathbf{h}_{u}^{(k-1)} \mid u \in \mathcal{N}(v) \right\} \right) \right),
   \]
   where \( \text{MEAN}(\cdot) \) denotes the mean aggregation function applied to the concatenated embeddings of node \( v \) and its neighbors.
\end{enumerate}   
GraphSAGE performs these processes iteratively across multiple layers (or hops). In each layer, nodes aggregate information from neighbors sampled in the previous layer, allowing the model to capture information from increasingly distant parts of the graph.
This process enables GraphSAGE to effectively learn node embeddings that encode both local and broader structural information from the graph.

In our case, we use two SAGEConv layers for training each graph so that each node in the graph can aggregate information not just from immediate neighbors (1-hop neighbors) but also from neighbors of its neighbors (2-hop neighbors), thus incorporating more contextual information from the graph. The \( \mathrm{ReLU}(\cdot) \) activation function applied after the first SAGEConv layer introduces non-linearity, allowing the network to learn more complex mappings. The output node representations of the four graphs are then concatenated before applying the ReLU activation function and feeding into a linear layer to get the demand prediction for each OD pair on the graph.

Let \( G_{i} \) denote the \( i \)-th graph where \( i \in \{t, s, o, d\} \) and \( F_{i} \) be the input node features for graph \( G_{i} \). The calculation procedure is shown in Equations \ref{eq5}-\ref{eq10}.

\begin{align}
    &H_i^{(1)} = \mathrm{SAGEConv}_1(F_i, A^i), \quad \text{for } i \in \{t, s, o, d\} \label{eq5}\\ 
    &H_i^{(1, \mathrm{ReLU})} = \mathrm{ReLU}(H_i^{(1)}), \quad \text{for } i \in \{t, s, o, d\} \label{eq6}\\
    &H_i^{(2)} = \mathrm{SAGEConv}_2(H_i^{(1, \mathrm{ReLU})}, A_i), \quad \text{for } i \in \{t,s,o,d\} \label{eq7}\\
    &H^{(\mathrm{Concat})} = \mathrm{Concat}(H_t^{(2)}, H_s^{(2)}, H_o^{(2)}, H_d^{(2)}) \label{eq8}\\
    &H^{(\mathrm{Concat, \mathrm{ReLU})}} = \mathrm{ReLU}(H^{(\mathrm{Concat})}) \label{eq9}\\
    &\hat{Y} = \mathrm{Linear}(H^{(\mathrm{Concat, \mathrm{ReLU})}}) \label{eq10}
\end{align}

Here, $H_i$ represent the feature representation of nodes in graph $G_{i}$ and $\hat{Y}$ is the predicted demand for all OD in urban rail.

\section{Computational Experiments}
\label{experiment}
 In this section, we present our experiments conducted for evaluating the proposed method with a real-world case study on Copenhagen S-train URT. 

\subsection{Data description}
We introduce a real-world case study for short-term OD prediction in URT based on data from the Copenhagen suburban railway, known as the S-train. The railway network consists of 170 km of electrified double track with homogeneous passenger traffic. The OD-demand dataset is constructed based on the Danish nationwide Automatic Fare Collection (AFC) system called 'Rejsekort,' a smart card system where users tap in (origin) and tap out (destination) are recorded. We collect demand information over 84 stations in the railway network over an 11-month period from January 29, 2021, to December 3, 2021. Data are collected only 5 days per week during peak periods from 5 AM to 12 PM. We adjust for holidays by using the historical average demand from the week before each holiday. The final demand data encompass tap-in and tap-out at all stations in the rail network (84 stations), which are then used to prepare the aggregated demand feature $\textbf{X}$ and completed trips of all ODs $\textbf{D}$ with a time interval of $t = 20$ minutes. We note that this AFC dataset does not account for all demand in the network, but it represents the main demand patterns on the Danish public transportation network \cite{eltved2020modelling}. Future work is required to include additional data to scale up the current framework to the total demand.

In addition to the tap-in and tap-out demand data, we also collect data on the number of trains, train delay times, and the number of trains canceled at each station per hour for each line and direction, which are then used to prepare supply features $\textbf{f}_{s}$. The location data of the 84 stations is obtained via the open API of Rejseplanen, a website providing Denmark's public transport timetable and information. For the training and test set split, we randomly choose 60 days within the 11-month period as test data, while the remaining days serve as training data.
% \begin{figure}[H]
%     \centering
%     \includegraphics[width = 0.6\textwidth]{Fig/OD_Visuzalization.png}
%     \caption{ODs map in Copenhagen}
%     \label{fig:OD}
% \end{figure}

\subsection{Experimental setup and baseline methods}
\begin{figure}[H]  
     \centering
     \begin{subfigure}[b]{0.48\textwidth}
         \centering
         \includegraphics[width=\textwidth]{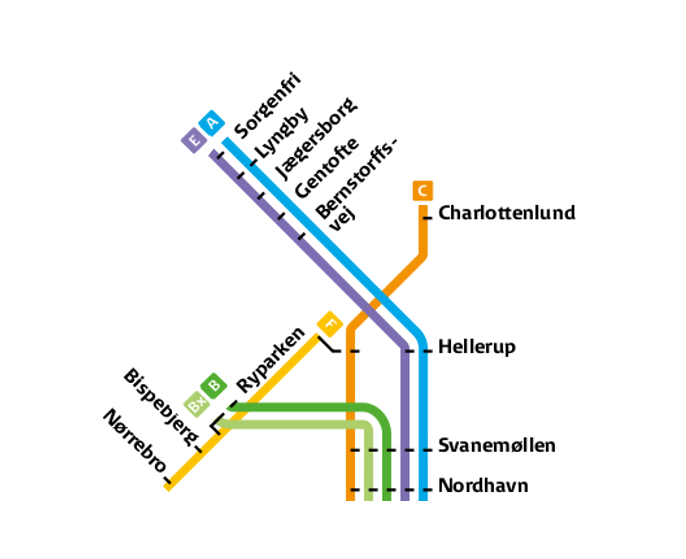}
         \caption{Tiny Copenhagen case}
         \label{f31}
    \end{subfigure}  
     \begin{subfigure}[b]{0.48\textwidth}
         \centering
         \includegraphics[width=\textwidth]{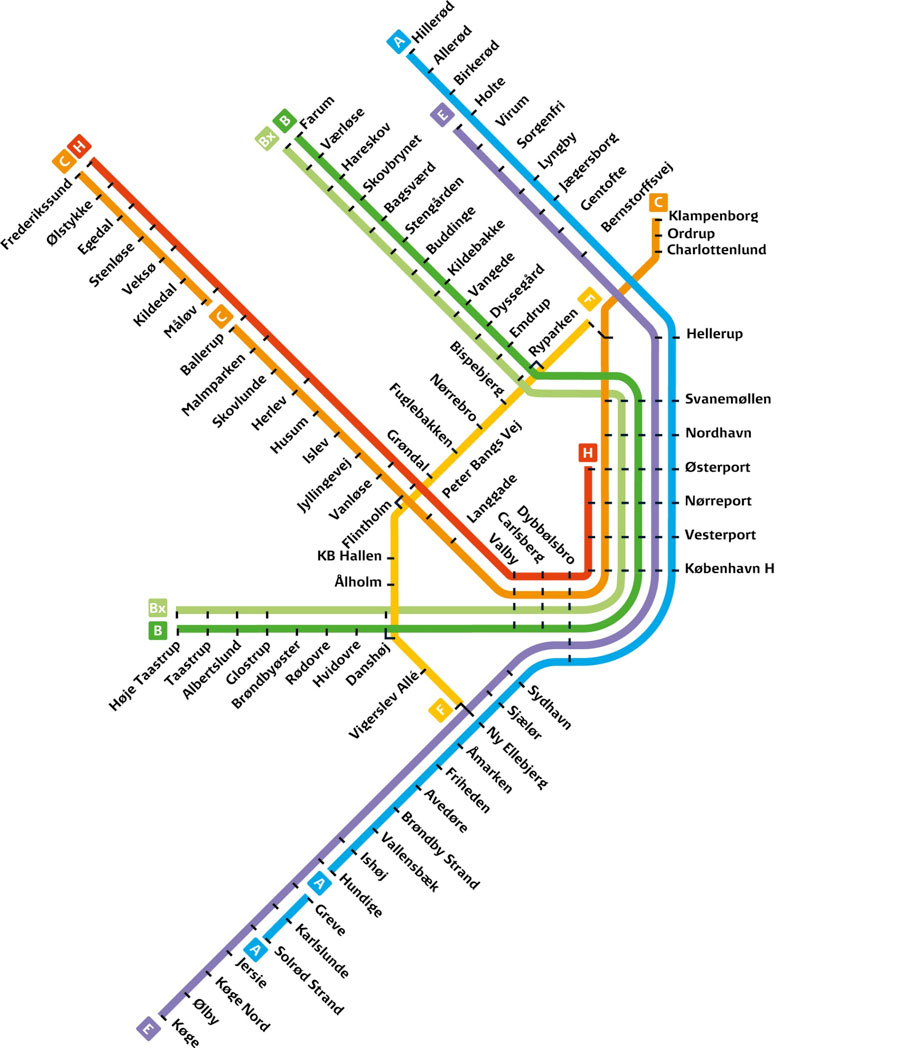}
         \caption{Full Copenhagen case}
         \label{f32}
     \end{subfigure}
        \caption{URT line maps of real-world S-train case studies}
        \label{f3}
%%\end{figure*}
\end{figure}

We evaluate our proposed method in two dimensions: scalability and robustness. For the scalability dimension, we test our method on three different graph scales of the railway network, including the "12 ODs" case, the "Tiny Copenhagen" case, and the "Full Copenhagen" case. In the 12 ODs case, we select the 12 OD pairs with the highest average demand. The "Tiny Copenhagen" case encompasses 12 contiguous stations in the network, resulting in 132 OD pairs, while the "Full Copenhagen" case covers the entire railway network with 84 stations, resulting in 6972 OD pairs (see Figure \ref{f3}).

%%\begin{figure*}

For comparative analysis, we employ different prediction methods as baselines as follows:
\begin{itemize}
\item \textbf{RR}: Ridge Regression is a type of linear regression that includes a regularization term to prevent overfitting, especially when dealing with multicollinearity or when the number of predictors is larger than the number of observations \cite{rodrigues2023importance}. The regularization parameter $\lambda$ is tuned from 1e-3 to 1e-7, with the optimal value being 1e-6.
\item \textbf{XGB}: XGBoost is an advanced implementation of gradient-boosted decision trees designed to be highly efficient, flexible, and portable \cite{DBLP:journals/corr/ChenG16}. XGBoost provides a significant improvement over traditional gradient-boosting algorithms for different classification and regression predictive modeling problems. The learning rate is tuned among 0.1, 0.01, and 0.001, while the maximum depth of the tree is tuned to 2, 4, and 6. For the 12 ODs case, the optimal learning rate is 0.1, while for the Tiny and Full Copenhagen cases, the optimal learning rate is 0.001. The best parameter for maximum depth is 6.
\item \textbf{GCN}: Graph Convolutional Neural Network is a graph representation learning method that serves as the foundation for many OD prediction models. Our GCN model contains two layers with 64 hidden units and a learning rate of 1e-4. GCN extends the concepts of convolutional neural networks (CNN) to graph-structured data.

% \item \textbf{NRI-GNN}: Neural Relational Inference GNN model is a state-of-the-art prediction model based on GNN \cite{tygesen2023unboxing}. In NRI-GNN, the graph adjacency matrix is learned by the encoder based on the NRI method \cite{kipf2018neuralrelationalinferenceinteracting}, while the decoder is a Gated Recurrent Unit (GRU) that takes the features and the dynamic graph generated by the encoder as input and outputs the predicted demand at time $t+1$ for all nodes. In our experiment, NRI-GNN is used as a baseline to evaluate the performance of the proposed method on the 12 ODs case and the Tiny Copenhagen case. For the Full Copenhagen case, NRI-GNN runs out of memory and is not scalable.  %Additionally, NRI-GNN needs weather information as input features for the model. 

\end{itemize}
To address the Full Copenhagen case and manage a large-scale graph within a reasonable computing and training time, we propose several alternatives. Specifically, to formulate a temporal correlation-based graph, we use the Fast Fourier Transform (FFT) instead of Dynamic Time Warping (DTW) to measure the similarity of demand time series data between OD pairs. DTW has a time complexity of \(O(n^2)\), which is significantly more time-consuming compared to FFT's \(O(n \log n)\). For the origin-destination distance graph, instead of setting a threshold to establish connections between OD pairs, we limit the number of edges to 10,000. This limit is determined empirically through experiments to ensure an appropriate training time. This helps avoid increasing graph complexity, which could slow down the training process. 

% We evaluate the prediction error of all models using two metrics: RMSE (Root Mean Square Error) and MAE (Mean Absolute Error).

We evaluate the prediction error of all models using two metrics: RMSE (Root Mean Square Error) and MAE (Mean Absolute Error). To assess the model’s performance under different reliability conditions of the URT system, we test the prediction model across periods with varying levels of train delays and cancellations, as follows:
\begin{itemize}
    \item All periods: the entire test time horizon. 
    \item Periods with an average number of train cancellations at the origin or destination station in the last hour greater than 0.
    \item Periods with an average train delay time at the origin or destination station in the last hour greater than 60, 180, and 300 seconds.
\end{itemize}

% However, evaluating the model under different scales and different reliability-related states of the system helps to understand the specific characteristics of the prediction model that are appropriate for dealing with real-world prediction tasks, beyond just focusing on prediction accuracy. Thus, we also test the prediction model by selecting periods when unexpected events occur in the URT system, such as cancellations or delays. More specifically, the different prediction periods for model robustness evaluation are as follows:
% \begin{itemize}
%     \item Periods with an average number of train cancellations at the origin/destination station in the last hour greater than 0. 
%     \item Periods with an average train delay time at the origin/destination station in the last hour greater than 60, 180 and 300 seconds.
% \end{itemize}

Finally, all the models are implemented and executed with PyTorch on a desktop computer with an AMD Ryzen Threadripper 3960X 3.8GHz CPU, 128 GB of RAM, and an NVIDIA GeForce RTX 3080 Ti GPU.

\subsection{Results}
The average prediction error for baseline methods compared to mGraphSAGE across the 12 ODs, Tiny Copenhagen, and Full Copenhagen cases is presented in Tables \ref{tab1}, \ref{tab2}, and \ref{tab3}. Bold values indicate the best prediction performance. Training times for mGraphSAGE are 1 hour, 5 hours, and 48 hours for these three network scales, respectively. We also consider the prediction error in different scenarios involving uncertainties in the URT system, as previously described.\\
% Please add the following required packages to your document preamble:
% \usepackage{graphicx}
\begin{table}[H]
\caption{Results for the 12 ODs case}
\label{tab1}
\resizebox{\textwidth}{!}{%
\begin{tabular}{lcccccccc}
\toprule
Methods                                               & \multicolumn{2}{c}{LR} & \multicolumn{2}{c}{XGB}         & \multicolumn{2}{c}{GCN} & \multicolumn{2}{c}{mGraphSAGE} \\
Metrics                                               & RMSE       & MAE       & RMSE           & MAE            & RMSE       & MAE        & RMSE      & MAE                \\
\midrule
All periods                                           & 3.226      & 2.119     & \textbf{3.179} & \textbf{2.095} & 3.254      & 2.152      & 3.189     & 2.097              \\
Cancelations at origin \textgreater 0 period          & 3.458      & 2.561     & \textbf{3.365} & 2.519          & 3.475      & 2.600      & 3.368     & \textbf{2.518}     \\
Cancelations at destination \textgreater 0 period     & 3.382      & 2.483     & \textbf{3.275} & \textbf{2.432} & 3.388      & 2.509      & 3.294     & 2.440              \\
Mean delays at origin \textgreater 60 periods         & 3.067      & 2.086     & \textbf{3.016} & \textbf{2.058} & 3.118      & 2.139      & 3.032     & 2.069              \\
Mean delays at origin \textgreater 180 periods        & 2.292      & 1.849     & \textbf{2.271} & \textbf{1.843} & 2.372      & 1.927      & 2.312     & 1.870              \\
Mean delays at origin \textgreater 300 periods        & 2.209      & 1.799     & \textbf{2.120} & \textbf{1.744} & 2.253      & 1.852      & 2.203     & 1.799              \\
Mean delays at destination \textgreater 60   periods  & 2.933      & 1.976     & \textbf{2.866} & \textbf{1.983} & 2.941      & 2.003      & 2.901     & 1.954              \\
Mean delays at destination \textgreater 180   periods & 2.179      & 1.706     & \textbf{2.148} & \textbf{1.685} & 2.248      & 1.760      & 2.197     & 1.714              \\
Mean delays at destination \textgreater 300   periods & 2.154      & 1.828     & \textbf{2.044} & \textbf{1.754} & 2.193      & 1.888      & 2.146     & 1.845       \\
\bottomrule
\end{tabular}%
}
\end{table}\textbf{}
% Please add the following required packages to your document preamble:
% \usepackage{graphicx}
\begin{table}[H]
\caption{Results for the  Tiny Copenhagen case}
\label{tab2}
\resizebox{\textwidth}{!}{%
\begin{tabular}{lcccccccccc}
\toprule
Methods                                               & \multicolumn{2}{c}{RR} & \multicolumn{2}{c}{XGB} & \multicolumn{2}{c}{GCN}  & \multicolumn{2}{c}{mGraphSAGE} \\
Metrics                                               & RMSE       & MAE       & RMSE       & MAE        & RMSE       & MAE          & RMSE           & MAE           \\
\midrule
All periods                                           & 0.712      & 0.439     & 0.712      & 0.438      & 0.700      & 0.427         & \textbf{0.688}          & \textbf{0.426}         \\
Cancelations at origin \textgreater 0          & 0.737      & 0.499     & 0.735      & 0.497      & 0.728      & 0.486            & \textbf{0.718}          & \textbf{0.484}         \\
Cancelations at destination \textgreater 0     & 0.701      & 0.486     & 0.704      & 0.487      & 0.696      & 0.476         & \textbf{0.683}          & \textbf{0.474}         \\
Mean delays at origin \textgreater 60s         & 0.727      & 0.466     & 0.732      & 0.469      & 0.716      & 0.454      & \textbf{0.700}          & \textbf{0.450}         \\
Mean delays at origin \textgreater 180s        & 0.713      & 0.468     & 0.722      & 0.471      & 0.702      & 0.454     & \textbf{0.687}          & \textbf{0.453}         \\
Mean delays at origin \textgreater 300s        & 0.796      & 0.530     & 0.805      & 0.528      & 0.785      & \textbf{0.511}        & \textbf{0.771}          & 0.514         \\
Mean delays at destination \textgreater 60s  & 0.751      & 0.484     & 0.760      & 0.488      & 0.743      & 0.476         & \textbf{0.723}          & \textbf{0.468}         \\
Mean delays at destination \textgreater 180s & 0.767      & 0.502     & 0.775      & 0.507      & 0.751      & 0.489     & \textbf{0.737}          & \textbf{0.490}         \\
Mean delays at destination \textgreater 300s & 0.767      & 0.528     & 0.772      & 0.528      & 0.750      & 0.512   & \textbf{0.737}          & \textbf{0.516}   \\
\bottomrule
\end{tabular}%
}
\end{table}
% Please add the following required packages to your document preamble:
% \usepackage{graphicx}
\begin{table}[H]
\caption{Results for the Full Copenhagen case}
\label{tab3}
\resizebox{0.9\textwidth}{!}{%
\begin{tabular}{lcccccc}
\toprule
Methods                                               & \multicolumn{2}{c}{RR} & \multicolumn{2}{c}{XGB} & \multicolumn{2}{c}{mGraphSAGE} \\
Metrics                                               & RMSE       & MAE       & RMSE       & MAE        & RMSE           & MAE           \\
\midrule
All periods                                           & 0.345      & 0.186     & 0.338      & 0.182      & \textbf{0.332}          & \textbf{0.181}         \\
Cancelations at origin \textgreater 0          & 0.257      & 0.180      & 0.256      & 0.180      & \textbf{0.247}          & \textbf{0.173}         \\
Cancelations at destination \textgreater 0    & 0.256      & 0.177     & 0.254      & 0.177      & \textbf{0.246}          & \textbf{0.17}          \\
Mean delays at origin \textgreater 60s         & 0.255      & 0.163     & 0.250      & 0.160      & \textbf{0.246}          & \textbf{0.158}         \\
Mean delays at origin \textgreater 180s        & 0.198      & 0.157     & 0.199      & 0.159      & \textbf{0.194}          & \textbf{0.154}         \\
Mean delays at origin \textgreater 300s        & 0.230       & 0.187     & 0.227      & 0.183      & \textbf{0.221}          & \textbf{0.178}         \\
Mean delays at destination \textgreater 60s  & 0.239      & 0.158     & \textbf{0.234}      & \textbf{0.155}      & 0.237          & 0.159         \\
Mean delays at destination \textgreater 180s & 0.321      & 0.238     & 0.326      & 0.242      & \textbf{0.316}          & \textbf{0.236}         \\
Mean delays at destination \textgreater 300s & 0.479      & 0.376     & 0.484      & 0.380      & \textbf{0.471}          & \textbf{0.370}         \\
\bottomrule
\end{tabular}%
}
\end{table}

Our results indicate that classic models remain competitive compared to the complex and computationally demanding graph-learning-based approaches, especially as the studied network (and its associated data) scales up. Nevertheless, informative graphs still prove valuable in capturing dependencies in the data, particularly spatial and temporal correlations often influenced by operations.

In the case of 12 ODs, the performance of mGraphSAGE is not particularly impressive. It is slightly better than RR and exhibits relatively similar performance to XGB. In some periods, especially when train delays exceed 180 seconds, mGraphSAGE performs worse than RR. Meanwhile, another GNN-based method, GCN, performs the worst among all methods. The underwhelming performance of GNN-based methods (mGraphSAGE and GCN) can be attributed to the selection of the 12 ODs for this experiment, which was based on demand criteria. These selected OD pairs exhibit little to no spatial correlation, diminishing the effectiveness of GNN-based methods in leveraging information from neighboring ODs to predict future demand. Additionally, evaluating the performance of methods based on the error in predicting demand for 12 ODs is more susceptible to bias, as poor prediction performance for a single OD pair can significantly impact the overall prediction error.

The differences in performance become more apparent when focusing on a specific part of the URT network—the Tiny Copenhagen case—where mGraphSAGE outperforms all other methods. In this scenario, RR and XGB show comparable performance, while GCN slightly outperforms these two traditional machine learning approaches. This demonstrates that when spatial correlation plays a role in prediction, GNN-based methods can effectively utilize the connections among nodes (OD pairs) on the graph to leverage information from neighboring nodes for better future demand predictions. Additionally, during specific periods marked by train delays and cancellations, the gap in prediction error between mGraphSAGE and other methods widens compared to predictions over all periods. For example, across all periods, mGraphSAGE achieves a 3.49\% lower prediction error than XGB. However, during periods when the mean delays at destination stations exceed 180 seconds, this difference increases to 5.16\%. This highlights mGraphSAGE's ability to effectively utilize features of neighboring nodes for predictions, especially under disruptions in the URT network.

When scaling the graph to the Full Copenhagen case with 6972 ODs, GCN fails to handle the large graph due to scalability issues and memory constraints. GCN requires loading the entire graph into memory for matrix operations, making it computationally expensive and memory-intensive for large graphs. In contrast, mGraphSAGE uses a sampling-based aggregation mechanism, where it samples a fixed number of neighbors for aggregation instead of considering all neighbors. This approach makes mGraphSAGE significantly more scalable and efficient for large graphs by reducing computational overhead. In this case, mGraphSAGE clearly outperforms the two traditional machine learning methods (RR and XGB), even with some customizations implemented to reduce computational effort compared to the Tiny Copenhagen case, such as using FFT to measure time correlations and relaxing node-ID features. Although the gap in prediction error between mGraphSAGE and XGB is not as large as in the Tiny Copenhagen case, the performance difference remains around 2–3\%, depending on the disruption scenarios. However, in cases where the mean delays at destination stations exceed 60 seconds, mGraphSAGE performs slightly worse than XGB. Future work could further improve the performance of mGraphSAGE through more carefully designed customizations.

Additionally, it is worth noting that there are many OD pairs with zero demand during several specific periods. Therefore, when averaging prediction error over a large number of OD pairs, even small differences in average prediction error have a significant impact on accuracy. To support this argument, we conducted statistical tests to verify the improvement of mGraphSAGE in prediction performance on all OD pairs compared to other methods for all scenarios in Tiny and Full Copenhagen cases. The p-values of the t-statistic comparing mGraphSAGE with other models are all less than 0.05.

\section{Conclusion}
\label{conclude}
In this paper, we propose a prediction model named mGraphSAGE for OD demand prediction in large-scale URT networks under operational uncertainties. Our method models each OD pair as a node on a graph and establishes different connectivities based on spatial and temporal correlations of OD pairs, resulting in a multiple graph framework. We then employ graph inductive representation learning to learn node representations from multiple graphs, enabling prediction of future OD demand at the network level. 

Experiments are conducted on three scenarios of varying scale using data from the S-train in Copenhagen, Denmark. Comparisons with different machine learning and GNN-based methods demonstrate that mGraphSAGE outperforms other methods in medium and large graphs, and can effectively handle real-world cases with 6972 ODs, where other GNN-based methods are not scalable. We also tested the prediction model under specific periods with unexpected operational issues, such as train delays and cancellations. Results show that mGraphSAGE outperforms other methods due to its ability to leverage graph connections for prediction under disrupted conditions. 

We still note that the performance of mGraphSAGE declined in Full Copenhagen compared to Tiny Copenhagen due to necessary customizations to reduce time complexity and input dimensions, including changes to node-ID one-hot encoding features, replacement of DTW with FFT, and manual setting of the maximum number of edges in the graphs. Also, carefully built classical models can be advantageous since they are trivial to implement and train and may result in minor  prediction deterioration. Yet, as observed in the comparison between the Tiny and Full Copenhagen cases, when dealing with even more complex and unreliable networks with interdependent demand patterns, proper graph representation should have a greater benefit. Indeed, dynamic OD prediction is particularly relevant when disruptions occur in the system. When this happens, prediction may be used for driving traffic control strategies, to best fit the actual demand. In particular in the railway context, where the evolution of traffic control systems is currently a major research and development topic \cite{ERJU}, the integration of OD prediction and traffic control has been shown to bring service quality improvements under various perspectives. 

To further improve prediction accuracy in disrupted situations, future work on the proposed graph-based architectures should: (i) be tested on more complex networks, (ii) incorporate additional inputs for full demand observability, such as other demand data sources and weather variables, as previously mentioned, (iii) be evaluated using alternative and additional graph formulations, potentially informed by initial data analysis to support model architecture design, and (iv) explore the concept of learning a dynamic adjacency matrix during model training, which could further enhance model performance.

% %% The Appendices part is started with the command \appendix;
% %% appendix sections are then done as normal sections
% \appendix
% \section{Example Appendix Section}
% \label{app1}

% Appendix text.

% %% For citations use: 
% %%       \citet{<label>} ==> Lamport [21]
% %%       \citep{<label>} ==> [21]
% %%
% Example citation, See \citet{lamport94}.

%% If you have bib database file and want bibtex to generate the
%% bibitems, please use
%%
\bibliographystyle{elsarticle-num-names} 
\bibliography{cas-refs}

\begin{thebibliography}{22}
\expandafter\ifx\csname natexlab\endcsname\relax\def\natexlab#1{#1}\fi
\providecommand{\url}[1]{\texttt{#1}}
\providecommand{\href}[2]{#2}
\providecommand{\path}[1]{#1}
\providecommand{\DOIprefix}{doi:}
\providecommand{\ArXivprefix}{arXiv:}
\providecommand{\URLprefix}{URL: }
\providecommand{\Pubmedprefix}{pmid:}
\providecommand{\doi}[1]{\href{http://dx.doi.org/#1}{\path{#1}}}
\providecommand{\Pubmed}[1]{\href{pmid:#1}{\path{#1}}}
\providecommand{\bibinfo}[2]{#2}
\ifx\xfnm\relax \def\xfnm[#1]{\unskip,\space#1}\fi
%Type = Misc
\bibitem[{Denmark(2024)}]{Passenge59:online}
\bibinfo{author}{S.~Denmark}, \bibinfo{title}{Passengers and routes - statistics denmark}, \bibinfo{howpublished}{\url{https://www.dst.dk/en/Statistik/emner/transport/persontransport/passagerer-og-transportruter}}, \bibinfo{year}{2024}. \bibinfo{note}{(Accessed on 07/24/2024)}.
%Type = Article
\bibitem[{Jiang et~al.(2022)Jiang, Ma, and Koutsopoulos}]{jiang2022deep}
\bibinfo{author}{W.~Jiang}, \bibinfo{author}{Z.~Ma}, \bibinfo{author}{H.~N. Koutsopoulos},
\newblock \bibinfo{title}{Deep learning for short-term origin--destination passenger flow prediction under partial observability in urban railway systems},
\newblock \bibinfo{journal}{Neural Computing and Applications}  (\bibinfo{year}{2022}) \bibinfo{pages}{1--18}.
%Type = Misc
\bibitem[{Aboah et~al.(2021)Aboah, Johnson, and Shah}]{aboah2021identifyingfactorsinfluenceurban}
\bibinfo{author}{A.~Aboah}, \bibinfo{author}{L.~Johnson}, \bibinfo{author}{S.~Shah}, \bibinfo{title}{Identifying the factors that influence urban public transit demand}, \bibinfo{year}{2021}. \URLprefix \url{https://arxiv.org/abs/2111.09126}. \href{http://arxiv.org/abs/2111.09126}{{\tt arXiv:2111.09126}}.
%Type = Article
\bibitem[{Zhang et~al.(2021)Zhang, Che, Chen, Ma, and He}]{zhang2021short}
\bibinfo{author}{J.~Zhang}, \bibinfo{author}{H.~Che}, \bibinfo{author}{F.~Chen}, \bibinfo{author}{W.~Ma}, \bibinfo{author}{Z.~He},
\newblock \bibinfo{title}{Short-term origin-destination demand prediction in urban rail transit systems: A channel-wise attentive split-convolutional neural network method},
\newblock \bibinfo{journal}{Transportation Research Part C: Emerging Technologies} \bibinfo{volume}{124} (\bibinfo{year}{2021}) \bibinfo{pages}{102928}.
%Type = Article
\bibitem[{Noursalehi et~al.(2021)Noursalehi, Koutsopoulos, and Zhao}]{noursalehi2021dynamic}
\bibinfo{author}{P.~Noursalehi}, \bibinfo{author}{H.~N. Koutsopoulos}, \bibinfo{author}{J.~Zhao},
\newblock \bibinfo{title}{Dynamic origin-destination prediction in urban rail systems: A multi-resolution spatio-temporal deep learning approach},
\newblock \bibinfo{journal}{IEEE Transactions on Intelligent Transportation Systems} \bibinfo{volume}{23} (\bibinfo{year}{2021}) \bibinfo{pages}{5106--5115}.
%Type = Article
\bibitem[{Zhang et~al.(2024)Zhang, Zhang, Yang, Chen, Li, and Gao}]{zhang2024physics}
\bibinfo{author}{S.~Zhang}, \bibinfo{author}{J.~Zhang}, \bibinfo{author}{L.~Yang}, \bibinfo{author}{F.~Chen}, \bibinfo{author}{S.~Li}, \bibinfo{author}{Z.~Gao},
\newblock \bibinfo{title}{Physics guided deep learning-based model for short-term origin-destination demand prediction in urban rail transit systems under pandemic},
\newblock \bibinfo{journal}{Engineering}  (\bibinfo{year}{2024}).
%Type = Article
\bibitem[{Yu et~al.(2017)Yu, Yin, and Zhu}]{yu2017spatio}
\bibinfo{author}{B.~Yu}, \bibinfo{author}{H.~Yin}, \bibinfo{author}{Z.~Zhu},
\newblock \bibinfo{title}{Spatio-temporal graph convolutional networks: A deep learning framework for traffic forecasting},
\newblock \bibinfo{journal}{arXiv preprint arXiv:1709.04875}  (\bibinfo{year}{2017}).
%Type = Article
\bibitem[{Zhao et~al.(2019)Zhao, Song, Zhang, Liu, Wang, Lin, Deng, and Li}]{zhao2019t}
\bibinfo{author}{L.~Zhao}, \bibinfo{author}{Y.~Song}, \bibinfo{author}{C.~Zhang}, \bibinfo{author}{Y.~Liu}, \bibinfo{author}{P.~Wang}, \bibinfo{author}{T.~Lin}, \bibinfo{author}{M.~Deng}, \bibinfo{author}{H.~Li},
\newblock \bibinfo{title}{T-gcn: A temporal graph convolutional network for traffic prediction},
\newblock \bibinfo{journal}{IEEE transactions on intelligent transportation systems} \bibinfo{volume}{21} (\bibinfo{year}{2019}) \bibinfo{pages}{3848--3858}.
%Type = Article
\bibitem[{Ke et~al.(2021)Ke, Qin, Yang, Zheng, Zhu, and Ye}]{ke2021predicting}
\bibinfo{author}{J.~Ke}, \bibinfo{author}{X.~Qin}, \bibinfo{author}{H.~Yang}, \bibinfo{author}{Z.~Zheng}, \bibinfo{author}{Z.~Zhu}, \bibinfo{author}{J.~Ye},
\newblock \bibinfo{title}{Predicting origin-destination ride-sourcing demand with a spatio-temporal encoder-decoder residual multi-graph convolutional network},
\newblock \bibinfo{journal}{Transportation Research Part C: Emerging Technologies} \bibinfo{volume}{122} (\bibinfo{year}{2021}) \bibinfo{pages}{102858}.
%Type = Article
\bibitem[{Ali et~al.(2022)Ali, Zhu, and Zakarya}]{ali2022exploiting}
\bibinfo{author}{A.~Ali}, \bibinfo{author}{Y.~Zhu}, \bibinfo{author}{M.~Zakarya},
\newblock \bibinfo{title}{Exploiting dynamic spatio-temporal graph convolutional neural networks for citywide traffic flows prediction},
\newblock \bibinfo{journal}{Neural networks} \bibinfo{volume}{145} (\bibinfo{year}{2022}) \bibinfo{pages}{233--247}.
%Type = Misc
\bibitem[{Hamilton et~al.(2018)Hamilton, Ying, and Leskovec}]{hamilton2018inductiverepresentationlearninglarge}
\bibinfo{author}{W.~L. Hamilton}, \bibinfo{author}{R.~Ying}, \bibinfo{author}{J.~Leskovec}, \bibinfo{title}{Inductive representation learning on large graphs}, \bibinfo{year}{2018}. \URLprefix \url{https://arxiv.org/abs/1706.02216}. \href{http://arxiv.org/abs/1706.02216}{{\tt arXiv:1706.02216}}.
%Type = Article
\bibitem[{Liu et~al.(2022)Liu, Zhu, Li, Wu, Bai, and Lin}]{liu2022online}
\bibinfo{author}{L.~Liu}, \bibinfo{author}{Y.~Zhu}, \bibinfo{author}{G.~Li}, \bibinfo{author}{Z.~Wu}, \bibinfo{author}{L.~Bai}, \bibinfo{author}{L.~Lin},
\newblock \bibinfo{title}{Online metro origin-destination prediction via heterogeneous information aggregation},
\newblock \bibinfo{journal}{IEEE Transactions on Pattern Analysis and Machine Intelligence} \bibinfo{volume}{45} (\bibinfo{year}{2022}) \bibinfo{pages}{3574--3589}.
%Type = Article
\bibitem[{Zhu et~al.(2023)Zhu, Ding, Wei, Yi, Xu, and Wu}]{zhu2023two}
\bibinfo{author}{G.~Zhu}, \bibinfo{author}{J.~Ding}, \bibinfo{author}{Y.~Wei}, \bibinfo{author}{Y.~Yi}, \bibinfo{author}{S.~S.-D. Xu}, \bibinfo{author}{E.~Q. Wu},
\newblock \bibinfo{title}{Two-stage od flow prediction for emergency in urban rail transit},
\newblock \bibinfo{journal}{IEEE Transactions on Intelligent Transportation Systems} \bibinfo{volume}{25} (\bibinfo{year}{2023}) \bibinfo{pages}{920--928}.
%Type = Article
\bibitem[{Cheng et~al.(2022)Cheng, Tr{\'e}panier, and Sun}]{cheng2022real}
\bibinfo{author}{Z.~Cheng}, \bibinfo{author}{M.~Tr{\'e}panier}, \bibinfo{author}{L.~Sun},
\newblock \bibinfo{title}{Real-time forecasting of metro origin-destination matrices with high-order weighted dynamic mode decomposition},
\newblock \bibinfo{journal}{Transportation science} \bibinfo{volume}{56} (\bibinfo{year}{2022}) \bibinfo{pages}{904--918}.
%Type = Article
\bibitem[{Cats and Jenelius(2014)}]{cats2014dynamic}
\bibinfo{author}{O.~Cats}, \bibinfo{author}{E.~Jenelius},
\newblock \bibinfo{title}{Dynamic vulnerability analysis of public transport networks: mitigation effects of real-time information},
\newblock \bibinfo{journal}{Networks and Spatial Economics} \bibinfo{volume}{14} (\bibinfo{year}{2014}) \bibinfo{pages}{435--463}.
%Type = Misc
\bibitem[{Zhang et~al.(2017)Zhang, Zheng, and Qi}]{zhang2017deep}
\bibinfo{author}{J.~Zhang}, \bibinfo{author}{Y.~Zheng}, \bibinfo{author}{D.~Qi}, \bibinfo{title}{Deep spatio-temporal residual networks for citywide crowd flows prediction}, \bibinfo{year}{2017}. \URLprefix \url{https://arxiv.org/abs/1610.00081}. \href{http://arxiv.org/abs/1610.00081}{{\tt arXiv:1610.00081}}.
%Type = Article
\bibitem[{Kathuria et~al.(2020)Kathuria, Parida, and Sekhar}]{kathuria2020review}
\bibinfo{author}{A.~Kathuria}, \bibinfo{author}{M.~Parida}, \bibinfo{author}{C.~R. Sekhar},
\newblock \bibinfo{title}{A review of service reliability measures for public transportation systems},
\newblock \bibinfo{journal}{International Journal of Intelligent Transportation Systems Research} \bibinfo{volume}{18} (\bibinfo{year}{2020}) \bibinfo{pages}{243--255}.
%Type = Article
\bibitem[{Tirachini et~al.(2022)Tirachini, Godachevich, Cats, Mu{\~n}oz, and Soza-Parra}]{tirachini2022headway}
\bibinfo{author}{A.~Tirachini}, \bibinfo{author}{J.~Godachevich}, \bibinfo{author}{O.~Cats}, \bibinfo{author}{J.~C. Mu{\~n}oz}, \bibinfo{author}{J.~Soza-Parra},
\newblock \bibinfo{title}{Headway variability in public transport: A review of metrics, determinants, effects for quality of service and control strategies},
\newblock \bibinfo{journal}{Transport Reviews} \bibinfo{volume}{42} (\bibinfo{year}{2022}) \bibinfo{pages}{337--361}.
%Type = Phdthesis
\bibitem[{Eltved(2020)}]{eltved2020modelling}
\bibinfo{author}{M.~Eltved}, \bibinfo{title}{Modelling passenger behaviour in mixed scheduleand frequency-based public transport systems}, Ph.D. thesis, Technical University of Denmark, \bibinfo{year}{2020}.
%Type = Inproceedings
\bibitem[{Rodrigues(2023)}]{rodrigues2023importance}
\bibinfo{author}{F.~Rodrigues},
\newblock \bibinfo{title}{On the importance of stationarity, strong baselines and benchmarks in transport prediction problems},
\newblock in: \bibinfo{booktitle}{2023 IEEE 26th International Conference on Intelligent Transportation Systems (ITSC)}, \bibinfo{organization}{IEEE}, \bibinfo{year}{2023}, pp. \bibinfo{pages}{4927--4932}.
%Type = Article
\bibitem[{Chen and Guestrin(2016)}]{DBLP:journals/corr/ChenG16}
\bibinfo{author}{T.~Chen}, \bibinfo{author}{C.~Guestrin},
\newblock \bibinfo{title}{Xgboost: {A} scalable tree boosting system},
\newblock \bibinfo{journal}{CoRR} \bibinfo{volume}{abs/1603.02754} (\bibinfo{year}{2016}). \URLprefix \url{http://arxiv.org/abs/1603.02754}. \href{http://arxiv.org/abs/1603.02754}{{\tt arXiv:1603.02754}}.
%Type = Misc
\bibitem[{ERJU(2024)}]{ERJU}
\bibinfo{author}{ERJU}, \bibinfo{title}{Europe's rail}, \bibinfo{year}{2024}. \URLprefix \url{https://rail-research.europa.eu/}, \bibinfo{note}{accessed November 5th, 2024}.

\end{thebibliography}

%% else use the following coding to input the bibitems directly in the
%% TeX file.

%% Refer following link for more details about bibliography and citations.
%% https://en.wikibooks.org/wiki/LaTeX/Bibliography_Management

% \begin{thebibliography}{00}

% %% For authoryear reference style
% %% \bibitem[Author(year)]{label}
% %% Text of bibliographic item

% \bibitem[Lamport(1994)]{lamport94}
%   Leslie Lamport,
%   \textit{\LaTeX: a document preparation system},
%   Addison Wesley, Massachusetts,
%   2nd edition,
%   1994.

% \end{thebibliography}
\end{document}